# Ant Colony Optimization of Rough Set for HV Bushings Fault Detection

L.J. Mpanza*, T. Marwala**
Department of Electrical and Electronics Engineering,
University of Johannesburg
Johannesburg, South Africa
*ljmpanza@gmail.com,
**tmarwala@uj.ac.za

*Abstract*—In this paper we propose the optimization of Rough Set method using ant colony for Oil-impregnated paper bushing. Ant colony is used to discretize the training data set. The ant colony optimized rough set is compare to a rough set who's data is discretized using Equal frequency binning (EFB). Ant colony optimized rough set results show an improvement compared to the EFB. The ACO Rough Set has an accuracy 4% high than that of EFB Rough set. Rules generated are only a third for ACO compared to AFB. Although AOC takes longer to train, it proves to outperform EFB in all other respects.

## I. INTRODUCTION

Data mining is a process of extracting useful information from a data set. This information is then used to analyze and predict new data. Over the years, mining tools have been proposed, each with its merits and demerits. The challenge that arises from development of these proposals is choosing the methods that best suit your application.

Fault detection is concerned with the identification of a fault and finding its cause. These two concerns require a method that is able to classify data accurately and also be able to reason why it is classified as such. The high accuracy is required to improve the hit rate of the classifier to detect the fault with confidence and timely and avoid any failure which may lead to financial loss. Its also required to reduce the false alarm rate which may result in unnecessary downtime and subsequently financial loss. When a fault is correctly detected it is necessary to find the cause of the fault so that a solution can be recommended. Hence, the second requirement is a method that is transparent and allows for easy interpretation of result.

Training a data mining model requires data to be prepared for the purpose. Discretization is a major part of the data preparation process [1]. This results from a fact that different discretization methods affect the learning ability of a model. Therefore, selecting the discretization method is just as important as selecting a data mining method.

This paper proposes the use of Rough Set for bushing fault detection. It further proposed the use of Ant colony optimization for data discretization. The next section reviews the literature relevant to this subject. Section III introduces the data used in this experiment, and then a description of the rough set theory for fault detection is given. Following the rough set theory is a brief on ant colony optimization and how it is used to optimize the rough set model. Section IV details the experiment setup. Section V presents the results of the experiment. This paper is then concluded by a summary of findings and the discussion of possible future work.

## II. LITERATURE REVIEW

Various methods have been developed for oil-filled-paper bushings fault detection. These methods include Key Gas analysis on dissolve-gas [2,3] and other computational intelligence techniques [4,5,6,7]. The drawback of Key Gas analysis is that it is based on the ratio of developed gas. Hence it requires a priori information to make an informed decision about the status of the bushing. On the other hand, the computational methods proposed are able to classify the bushing data. They, however, also have a limitation as they only focus on the improving accuracy of classification and not the discovery of knowledge about the fault detected.

Since its discovery, rough set has found central attention as a rule induction tool for interpreting vague, ambiguous and incomplete data. Rough set is suitable in modeling methods that are used for both classification and interpretation due to its rule base nature. Rough set is usually used as a feature extraction and dimension reduction tool in conjunction with another computational intelligence method [8, 9]. It has been used successfully in fault detection of bushings [10]. Attempts have also been made to improve the classification accuracy and to reduce the number of rules. Crossingham and Marwala [11] experimented with the optimization techniques to discretize the training data to improve the performance of the rough set. Mpanza and Marwala [10] investigated the method of reducing the number of rules without trading-off the accuracy. The results obtained were found to be satisfactory.

Since classification accuracy and the number of rules are the cornerstone of optimizing rough set [12], we develop a model based on the foundation laid forth by [11] and [12]. Ant colony is proposed in this paper to discretize the data and this proposed method is compared to a more conventional discretization method, Equal frequency binning.

## III. SYSTEM FORMULATION

### A. Dissolved gas analysis

Dissolved gas analysis (DGA) is one of the most used techniques for detecting faults in oil-filled transformer bushing. DGA is a periodic experiment of extracting insulation oil from the bushing and using spectroscopy to analyze its gaseous composition. The proportions of the extracted gases are indicators of whether there is a fault or not. DGA data collected on regular basis aids in detection of any change in the gaseous content of the oil to detect early development of faults.

There are nine gases of interest in DGA. Hydrogen ($H_2$), methane ($CH_4$), ethylene ($C_2H_4$), ethane ($C_2H_6$), acetylene ($C_2H_2$), carbon monoxide (CO) and carbon dioxide ($CO_2$) are fault gases, while nitrogen ($N_2$) and Oxygen ($O_2$) are non-fault gases. These gases develop as a result of the cellulose paper that in immersed in the insulation oil. When the cellulose paper decomposes, H2, CH4 and CO are produced to signal a fault. The presence of simple hydrocarbons signals a fault due to electrical or thermal stress. These production rates are based on the IEEEc57.104 [13] and IEC60599 standard [3]. DGA data is used to model a fault detection method using Rough Set.

### B. Rough set theory

Rough Set Theory is a data mining tool. It was discovered by Pawlak as a tool to describe incomplete, imprecise and uncertain information by using approximation for classification. When data is represented in tabular form, each row represents the case data (object). The columns are divided into two types. The first set of columns is called the condition attribute (variable) and is used to describe the data objects. The second set is called the decision attribute (outcome) and is used to classify the data. The value of the decision attribute is usually binary, represented as 0 or 1 for belonging to a particular class or not, respectively.

In rough set terminology, this tabular representation is called an Information System (IS) and is described as a pair:

$$\Lambda = (U, A) \quad (1)$$

where $U$, a non-empty set of objects is called the universe and $A$ is a nun-empty set of attributes. Given $a \in A$, there exist a set $V_a$ of values such that

$$a : V \to V_a \quad (2)$$

In supervised learning there is an IS with decision attributes d which transform (1) into a Decision System (DS):

$$\Lambda = (U, A \cup \{d\}) \quad (3)$$

where $d$ is a condition attribute. $d$ has a value set $V_d = \{v_d^1, ..., v_d^{r(d)}\}$. In a binary decision system, $r(d) = 2$, which is the cardinality of the value set $V_d$.

If B is the subset of A ($B \subseteq A$), then there is a binary relation $IND(B)$ on $U$ called the indiscernibility relation. When two objects are indiscernible it means they can not be distinguished from one another. This relation is described by B-indiscernibility and is represented as:

$$(x, y) \in IND(B) \text{ iff } a(x) = a(y)$$
$$x, y \in U \quad (4)$$

and each indiscernibility class is called an elementary set.

A crisp set is the union of all elementary sets. The remaining objects are called rough sets. A DS is rough iff not all objects can be assigned to only one of the elementary sets.

Rough set allows for the definition of two set approximations. The lower approximation, which represents the objects that are completely contained in the set $X$ of interest, is denoted by $\underline{B}X$. The upper approximation includes the objects that are partially contained in set $X$ of interest and is denoted by $\overline{B}X$. $\underline{B}X$ and $\overline{B}X$ are represented mathematically as:

$$\underline{B}X = \{x \in U : B(x) \subseteq X\} \quad (3)$$
$$\overline{B}X = \{x \in U : B(x) \cap X \neq \varnothing\} \quad (4)$$

The region between the lower and upper approximations is called the boundary region and is represented as $BN_X = \overline{B}X - \underline{B}X$. This region represents the uncertain objects, and is fundamental to RS. If $BN_X$ is non-empty then the set is rough, else it is crisp. The objects that can be ruled out as being contained in set $X$ are represented by $U - \overline{B}X$.

For the objects in the boundary region, rough set uses a membership function to evaluate the plausibility of $x$ being contained in set $X$. The membership function is defined as:

$$\mu_A^X : U \to [0,1] \text{ and } \mu_A^X = \frac{|B(x) \cap X|}{B(x)} \quad (5)$$

Rules are extracted from the lower and upper approximations to generalize the input-output relationship. The rules are represented in the form: *if (condition) – then (decision)*. Theses rules are used to classify future novel objects.

### C. Ant colony otimisation

Ant colony is an optimization tool falling into the category of swarm intelligence. It is a meta-heuristic paradigm used to estimate solution for combinatorial optimization problem. As opposed to analytical algorithms which try to find solutions by exhaustive search, meta-heuristic are approximate algorithms used to find near optimal solution in reasonable computation time.

Inspired by the behavior of real ants, ACO was developed by Dorigo [14]. It takes advantage of the natural behavior of ants in search for food. Ants start searching at random, but as soon as the food source has been identified, and the quantity and the quality of the food evaluated, ants deposit a pheromone which sends a signal to the other ants. On the next search iteration, ants choose the trail with the most pheromone deposit, while depositing their own, by so doing, reinforcing the found trail.

A problem is an optimization problem if

$$P = (S, f) \quad (6)$$

Where $S$ is a search space, set of finite solutions and $f$ is an objective function such that:

$$f: S \to \Re^*, \quad (7)$$

assigns a cost value to the solution $S$. The objective of optimization is to find the cost value of solution $S^*$ such that $f(S) \leq f(S^*)$ in a reasonable amount of computation time.

ACO defines a number of concurrent artificial ants, $k$ and a pheromone model. Ants move in the system based on the pheromone trail parameter $T$, which has pheromone values $\tau$ associated with a solution. Ant also uses the solutions' attractiveness $N$ to select a candidate solution. At the end of each iteration (ant path completed), the solutions are evaluated and the pheromone trails and the attractiveness parameters are updated. These updates make the bases of the next iteration. The pheromone trail evaporation is used to demerit the solutions which are not selected as candidate solution. This process is repeated until a termination condition is met.

The ACO system is defined as follows:

- Attractiveness matrix $N_{ij}$ – The desirability of selecting the path between $i$ and $j$,
- Pheromone trail level $T_{ij}$ – A posteriori desirability of selecting the path between $i$ and $j$, based on the previous selection of the path.
- Probability of ant $k$ at node $i$ of selecting path $j$ as the next node is given by:

$$p_j^k(i) = \begin{cases} \dfrac{\tau_{ij}^\alpha \cdot \eta_{ij}^\beta}{\sum_{m \in N_i^k} \tau_{im}^\alpha \cdot \eta_{im}^\beta} & if \quad j \in N_i \\ 0 & otherwise \end{cases}, \quad (8)$$

where $N_i$ is the set of nodes accessible from $i$ and $\alpha$ and $\beta$ are tuning parameters

- The path traversed by ant $k$ is $T_k$ – The solution found by ant $k$.
- The cost function $L_x$ – Is the error incurred by choosing path $T_k$.
- Change in pheromone trail is give by:

$$\Delta \tau_{ij}^k = \begin{cases} Q/L_k & if (i,j) \in T_k \\ 0 & otherwise \end{cases}. \quad (9)$$

- Pheromone are updated by:

$$\tau_{ij} = (1-\rho).\tau_{ij} + \sum_{k=0}^{m} \Delta \tau_{ij}^k \quad (10)$$

where $\rho$ is the pheromone evaporation constant.

Performance of the ACO is based on the correct selection of $k$-number of ants, number of search iterations and the tuning factors, $\alpha$, $\beta$ and $\rho$.

### D. Discretization

Discretization is the process of transforming continuous variables into discrete variables. The problem of discretization is one of deciding how many split-points to choose and where to place them. Various methods have been proposed to solve this problem of discretization, each method with its merits and demerits.

Discretization is divided into two categories: supervised and unsupervised. Supervised discretization uses the class information of the training data set to optimize the selection and placement of the data cuts. The cuts are selected in a way that maximizes the purity of the intervals. Unsupervised discretization does not use the class information.

The selection of a discretization method influences the learning ability of a machine. Hence, it is important to evaluate each method and select the one that optimizes the learning process.

We proposed a supervised discretization method based on the ACO. This method is the compared to the Equal Frequency discretization

## IV. EXPERIMENT SETUP

Figure 1 shows the process involved in developing an ACO optimized RS model for Bushing fault detection.

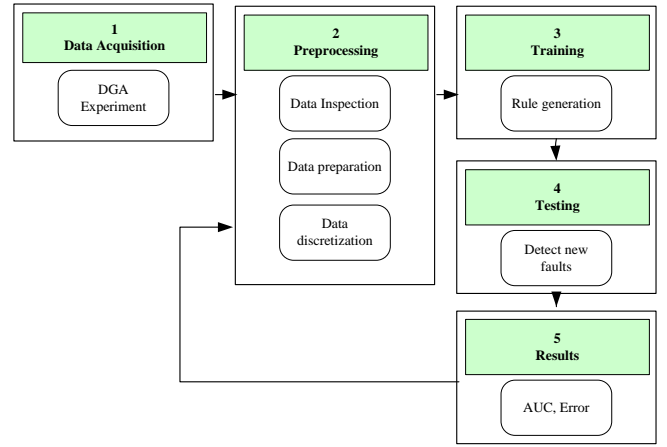

Figure 1. The development process of the ACO RS for fault detection

### A. DGA Experiement

Data is collected using DGA and saved in a form of a decision system (DI). Each experiment case is an object and the measured gases form the object's condition attributes. The decision attribute is added to the system and used to tell whether the object is faulty (1) or not (0).

### B. Data preprocessing

Preprocessing is divided into three parts: data inspection, data preparation and data discretization.

*1) Data inspection*

Data inspection is conducted to improve the quality of the data. This involves removal of outliers, solving missing value problem

*2) Data preparation*

Of th 60966 data set, 2000 was selected for the experiment. 1400 objects were assigned to the training set and 600 to the testing set.

*3) Data discretization*

Figure 2 shows the methodology employed in dicretizing a continuous variable

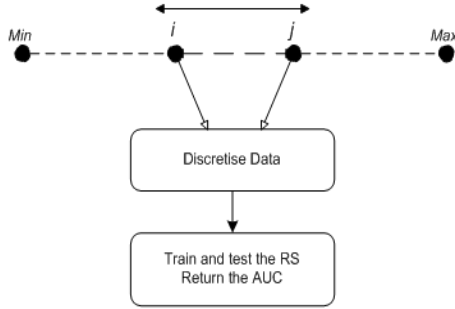

Figure 2. Depiction of discretizing a continuous variable. Min and Max represent the variable minimum and maximum, respectively. i and j are the selected cuts and can vary between Min and Max.

Two methods of discretization are compared to each other. The first is the unsupervised Equal Frequency Bin. We propose the second discretization method which uses ACO.

*a) Equal Frequency Bin (EFB)*

EFB is a univariate discretization technique that partitions a variable into predefined cuts such that all categories contain the same amount of object. For this experiment two cuts were chosen. Therefore, for each condition attribute, EFB selects i and j such that the intervals (Min-i), (i-j) and (j – Max) contain the same number of objects.

*b) Ant colony optimization (ACO)*

Conversely, the proposed method is a multivariate discretization model. ACO optimizes the placement of cuts by searching all the decision variable concurrently to minimize the error of the RS model being developed. Two cuts are used.

Figure 3 shows the ant colony flow chart. The process starts with the initialization of pheromone and attractiveness matrices. Then for each iteration, set an ant to a starting position. To make the search space countable, the i and j are allowed to assume only integer percentages of the continuous variable. Therefore $i \in [1,99]$ and $i < j < 100$. For each variable all ants start at *Min* and then compute the probability of moving to the next node $(i)$. This is done for all condition variables. When all ants are done selecting the solution, the cuts are used to discretize the training and the testing data. Using this data the RS is trained and tested. The classification error is used as the objective function. If the error is high, then cost of the solution is also high.

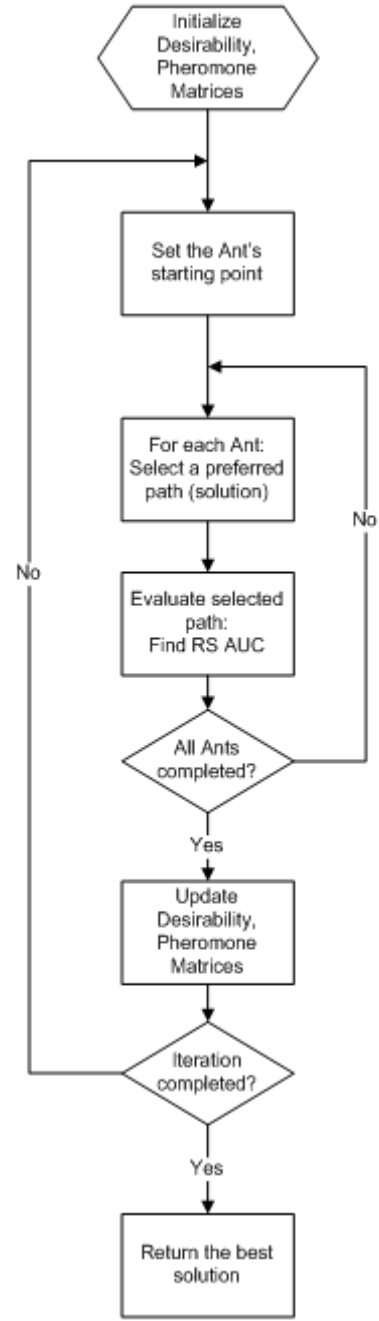

Figure 3. Ant colony flow chart - the process of determining the candidate solution for the placement of cuts

The path pheromone is updated using equation (5). This process is repeated until the number of iteration is reached. The pair that produced the best solution (least error) is returned as the final solution.

10 ants and 100 iterations are used. The tuning constants were chosen to be 0.09, 0.09 and 0.9 for [alpha], [beta] and [phi], respectively.

*4) Rule generation*

Using the data discretized by EFB and ACO, the RS is trained. The lower and upper approximations are constructed from the training data. Certain and non-certain rules are derived from the lower and upper approximations respectively.

*5) Solution evaluation*

To measure the effectiveness of the proposed solution against EFB a number of performance criteria are compared. There are: the nature of cuts, the time taken for the classifier to be developed, the number of rules generated and the classification accuracy. The latter is used to measure the ability of the classifier to correctly classify new and novel data. A confusion matrix is used to measure accuracy. Equation (11) uses confusion matrix information to calculate classification accuracy.

$$Accuracy = \frac{TP+TN}{TP+FP+FN+FN} \quad (11)$$

where $TP$, $TN$, $FP$ and $FN$ are True Positive, True Negative, False Positive and False Negative predictions respectively.

Receiver Operation Characteristic (ROC) curve is used to visualize the performance of the classifier, as it clearly shows the trade-offs between the correctly classified object and the false alarm. Area Under the Curve (AUC) is used to measure the generalization of the classifier. The higher the AUC, the better is the classifier in classifying unseen objects.

## V. RESULTS OBTAINED

TABLE I. THE CHARECTERISTICS OF THE EFB, ACO. THE TABLE SHOWS THE CONFUSION MATRIX, THE AREA UNDER THE CURVE (AUC), THE NUMBER OF RULES GENERATED AND THE DEVELOPMENT TIME (S) OF EACH CLASSIFIER

| Equal Frequency Bin | | | | | | |
|---|---|---|---|---|---|---|
|   | *PP* | *PN* | *AUC* | *# of Rules* | *Train Time(s)* | *Test Time(s)* |
| *AP* | 249 | 38 | 0.925 | 120 | 4 | <1 |
| *AN* | 39 | 274 | | | | |
| Ant Colony Optimized | | | | | | |
|   | *PP* | *PN* | *AUC* | *# of Rules* | *Time(s)* | *Test Time(s)* |
| *AP* | 222 | 69 | 0.961 | 45 | 723 | <1 |
| *AN* | 71 | 242 | | | | |

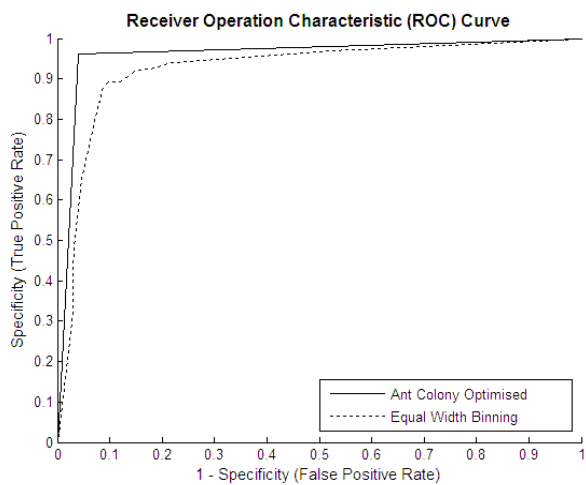

Figure 4. Receiver Operation Curve for Equal Frequency Bin and Ant Colony Optimised

Table I shows the characteristics of the developed methods. Figure 4 shows the generalization of both models on the ROC curve.

The confusion matrix shows the results from testing each classifier. Comparing the AUC and the number of rules generated, the ACO out performs the EFB. The only drawback with ACO is the training time. This is due to the fact that ACO has to search a number of candidate cuts before the final solution is created, while EFB is only executed once. This, however, poses no performance limitation on the AOC as it is a once off occurrence. Once the ACO model has been developed, it takes the same amount of time to classify objects.

## VI. CONCLUSION

This paper proposed a method of discretizing DGA for rough set using Ant Colony Optimization (ACO). This was compared with Equal frequency bin. ACO is modeled as a supervised discretization method using the class knowledge to partition data. It is also multivariate, since it uses the combination of all condition attributes in search for the best cuts. ACO proved to be superior to its unsupervised and univariate counterpart, EFB, in generalization, number of rules produced and accuracy. The ACO takes longer to train than EFB, but takes the same time to classify test objects.

For future research, more optimization methods can be compared to find the one that best represent the bushing fault detection. Also, the dimension reduction of the data and then optimizing the cuts selection of the resultant variable can be investigated. There is still a lot of room for improving the results found in this experiment.